\pdfoutput=1

\documentclass[11pt]{article}

\usepackage[final]{acl}

\usepackage{times}
\usepackage{latexsym}

\usepackage[T1]{fontenc}

\usepackage[utf8]{inputenc}

\usepackage{microtype}

\usepackage{inconsolata}

\usepackage{graphicx}
\usepackage{booktabs}

\title{IIITH-BUT system for IWSLT 2025 low-resource Bhojpuri to Hindi speech translation}

\author{
  \textbf{Bhavana Akkiraju\textsuperscript{1}},
  \textbf{Aishwarya Pothula\textsuperscript{1}},
  \textbf{Santosh Kesiraju\textsuperscript{2}},
  \textbf{Anil Kumar Vuppala\textsuperscript{1}} \\
  \textsuperscript{1}International Institute of Information Technology, Hyderabad, India \\
  \textsuperscript{2}Speech@FIT, Brno University of Technology, Czechia \\
  \texttt{\{bhavana.akkiraju,aishwarya.pothula\}@research.iiit.ac.in} \\
  \texttt{kesiraju@fit.vut.cz,anil.vuppala@iiit.ac.in}
}

\begin{document}
\maketitle
\begin{abstract}
This paper presents the submission of IIITH-BUT to the IWSLT 2025 shared task on speech translation for the low-resource Bhojpuri-Hindi language pair. We explored the impact of hyperparameter optimisation and data augmentation techniques on the performance of the SeamlessM4T model fine-tuned for this specific task. We systematically investigated a range of hyperparameters including learning rate schedules, number of update steps, warm-up steps, label smoothing, and batch sizes; and report their effect on translation quality. To address data scarcity, we applied speed perturbation and SpecAugment and studied their effect on translation quality. We also examined the use of cross-lingual signal through joint training with Marathi and Bhojpuri speech data. Our experiments reveal that careful selection of hyperparameters and the application of simple yet effective augmentation techniques significantly improve performance in low-resource settings. We also analysed the translation hypotheses to understand various kinds of errors that impacted the translation quality in terms of BLEU.
\end{abstract}

\section{Introduction}

Speech translation (ST) transforms spoken language into written text in a different language, serving as a critical component in breaking down communication barriers. While significant advancements have been made for high-resource language pairs \cite{jia2019direct, bentivogli2021cascade}, developing effective ST systems for low-resource and dialectal languages remains challenging due to scarce parallel data, inconsistent orthography, and substantial linguistic variation.

We address the low-resource scenario of Bhojpuri speech to Hindi text translation within the Indian linguistic context. Despite being spoken by over 50 million people, Bhojpuri suffers from limited quality and diversity in available speech-text corpora \cite{enwiki:1289695299}. In contrast, Hindi possesses relatively abundant resources, making it an ideal target language for translation. Our system, developed for the IWSLT 2025 shared task, addresses these resource disparities through a combination of transfer learning from linguistically similar languages, data augmentation techniques, and hyperparameter optimization.

Recent end-to-end ST models, such as the Speech-to-Text Transformer \cite{wang2020fairseq} and SeamlessM4T \cite{barrault2023seamlessm4t} effectively replace traditional cascade pipelines by jointly modelling ASR and MT, reducing error propagation and latency. However, these approaches typically require substantial labelled data unavailable for most low-resource languages, often leading to overfitting and poor generalisation.

Our contributions include:
\begin{itemize}
    \item A systematic investigation of hyperparameter optimization for low-resource ST, identifying that configuration choices such as batch sizes, moderate label smoothing values, and extended warmup periods significantly impact performance on the Bhojpuri-Hindi language pair.
    \item An analysis of data augmentation techniques - SpecAugment \cite{park2019specaugment} and speed perturbation \cite{ko2015audio} for low-resource speech translation, demonstrating their effectiveness in expanding our training data by 3x and improving BLEU scores by an average of 2.1 points.
    \item An evaluation of cross-lingual transfer learning through joint fine-tuning with Marathi-Hindi data, empirically showing how linguistic similarities between related Indo-Aryan languages can be leveraged to improve low-resource speech translation performance.
\end{itemize}

The remaining  of this paper is organized as follows: Section~\ref{sec:lit} provides a comprehensive review of related work, Section~\ref{sec:model} describes our system including hyperparameter optimization, data augmentation techniques, and joint fine-tuning approach, Section~\ref{sec:exp} details our experimental setup, Section~\ref{sec:results} presents results and analysis, and Section~\ref{sec:conc} concludes with future directions.

\section{Related work}
\label{sec:lit}
Low-resource speech translation (ST) has garnered significant attention through IWSLT shared tasks \cite{agrawal-etal-2023-findings,ahmad-etal-2024-findings,gow-smith-etal-2023-naver}. The field has evolved from traditional pipeline approaches \cite{post2013improved}  to end-to-end architectures such as Listen Attend and Spell (LAS) \cite{berard2016listen}, fairseq S2T \cite{wang-etal-2020-fairseq,e-ortega-etal-2023-quespa},\cite{radhakrishnan-etal-2023-sri}, and transfer learning based methods \cite{kesiraju23_interspeech,kesiraju-etal-2023-systems}. In \cite{mbuya-anastasopoulos-2023-gmu}, explored fine-tuning self-supervised models by incorporating a linear layer for the ST task, which streamlined workflows while maintaining specialized strategies for low-resource scenarios. \cite{vishnu-kudlu-shanbhogue-etal-2023-improving}, implemented various data augmentation techniques including audio stretching, back-translation, and paraphrasing.

Contemporary approaches to low-resource ST can be categorized into several methodological frameworks. The SETU-DCU submission \cite{zafar2024setu} enhanced ST robustness through CTC loss integration and rigorous data cleaning protocols. \cite{post2013improved} incorporated pseudo-labelling techniques to expand their training corpus. The JHU IWSLT 2024 system \cite{robinson2024jhu} demonstrated the efficacy of Whisper-style large models with domain-adaptive pretraining methodologies. Meanwhile, the QUESPA team \cite{ortega2024quespa} implemented ensemble decoding with cross-lingual knowledge transfer mechanisms. SeamlessM4T \cite{barrault2023seamlessm4t}  presents a comprehensive approach handling ASR, MT, and ST across more than 100 languages, though its performance on genuinely low-resource languages necessitates substantial adaptation strategies.

Our approach focuses on systematically exploring hyperparameter optimization and data augmentation techniques for low-resource speech translation. Unlike previous work that often applies general strategies such as default hyperparameter values, generic data augmentation, standard transfer learning without language-specific considerations and using default model architectures and training strategies without adaptation to low-resource constraints, we conduct a comprehensive investigation specifically tailored to the challenges of the Bhojpuri-Hindi language pair. Our experimentation includes a detailed analysis of learning rates, batch sizes, label smoothing values, and warmup periods, and  used two data augmentation techniques (Speed Perturb, SpecAug). Additionally, we examine how cross-lingual transfer from Marathi can supplement these optimizations, leveraging the linguistic proximity between these Indo-Aryan languages.

\section{Methodology}
\label{sec:model}
\subsection{Model architecture}
Our experiments used SeamlessM4T as the backbone. We experimented with medium (1.2B parameters) and large (2.3 B parameters) variants. The medium consists of a 24-layer conformer speech encoder, 12-layer Transformer text decoder, with 1024-dimensional hidden states, and 16 attention heads. 

\subsection{Fine-tuning}
For most of the experiments, we fine-tuned all the parameters on the target language pair, i.e. Bhojpuri--Hindi. We also conducted experiments where fine-tuned on both language pairs Marathi--Hindi and Bhojpuri--Hindi. This model was further fine-tuned for few epochs for the target pair Bhojpuri--Hindi.

\subsection{Evaluation}
We used the standard objective metrics BLEU \cite{papineni-etal-2002-bleu} and chrF++ \cite{popovic-2017-chrf}  implemented in sacrebleu \cite{post-2018-call}  to objectively evaluate the translation quality against the reference. 

\section{Experiments}
\label{sec:exp}

\subsection{Datasets}
We used only the official IWSLT 2025 shared task dataset for Bhojpuri → Hindi speech translation:
\begin{table}[ht]
\centering
\begin{tabular}{lrr}
\toprule
\textbf{Dataset} & \textbf{Duration (hrs)} & \textbf{No. of utterances} \\
\midrule
Training  & 20.00 & 10,171 \\
Dev & 2.07 & 1,056 \\
Test & 0.87 & 750 \\
\bottomrule
\end{tabular}
\caption{Data split statistics for Bhojpuri-Hindi ST task.}
\label{tab:data_stats}
\end{table}
For our multilingual experiments, we incorporated an additional Marathi $\rightarrow$ Hindi\footnote{\href{https://github.com/panlingua/iwslt2025_mr-hi}{IWSLT Marathi-Hindi Dataset}} parallel corpus from IWSLT, consisting of 16 hours (7,990 utterances) of training data, 3.6 hours (2,103 utterances) of development data, and 0.45 hours (286 utterances) of test data.

\subsection{Implementation}
Our code was based on the original seamless library written in PyTorch.
Our experiments used two NVIDIA A100 GPUs (40GB each), employing data parallel training, synchronised batch normalisation, and FP16 mixed precision for optimal computational efficiency.

\subsection{Pre-trained model selection}
Here we report the fine-tuning results for Seamless medium and large v2 variants. We observed that the medium model consistently yielded better translation results than the large one. Hence, we used the medium variant for all subsequent experiments and analysis.

\begin{table}[!ht]
    \centering
    \begin{tabular}{lrrr}
    \toprule
    \textbf{Model} & lr & BLEU & chrF++ \\ \midrule
     Medium & 1e-5 & \textbf{30.5} & \textbf{54.6} \\
     Large  & 1e-5  &25.5 & 50.5\\
     
     \bottomrule
    \end{tabular}
    \caption{Translation scores on dev set after fine-tuning seamless medium and large variants on  Bhojpuri--Hindi.}
    \label{tab:med_vs_large}
\end{table}

\subsection{Hyperparameter optimisation}
Our hyperparameter optimisation investigation focused on several critical training parameters. The \textbf{learning rate} (LR) was evaluated across three settings (1e-6, 1e-5, and 2e-5), with the moderate rate of \textbf{1e-5} consistently yielding the best performance as shown in Table~\ref{tab:lr_comp}, achieving a balance between training stability and domain adaptation. For \textbf{label smoothing} (LS), we tested values of 0, 0.1, and 0.2, with \textbf{0.1} offering the best generalisation and robustness (Table~\ref{tab:ls_comp}). We explored \textbf{warmup steps} (100, 250, 350, and 400) to stabilise the learning rate schedule, finding \textbf{250 steps} produced optimal convergence without extending training time. To mitigate overfitting, we experimented with \textbf{early stopping patience} values (5, 10, and 20 epochs), where \textbf{10 epochs} struck the best trade-off between overtraining and early termination. Finally, we examined \textbf{training batch size} (5, 10, 32, and 64), with a batch size of \textbf{32} demonstrating the most favourable performance (Table~\ref{tab:lr_comp}), balancing gradient update stability with computational efficiency.
\begin{table}[th]
  \centering
  \begin{tabular}{lrrr}
    \toprule
    \textbf{LR} & \textbf{batch size} & \textbf{BLEU} & \textbf{chrF++} \\
    \midrule
    1e-5 & 5   & 30.5 & 54.5 \\
    1e-5 & 10  & 33.8 & 56.9 \\
    1e-5 & 32 & \textbf{35.1} & 58.2 \\ 
    \hline
    1e-6 & 5   & 18.2 & 48.4 \\
    1e-6 & 10  & 20.1 & 49.1 \\
    1e-6 & 32  & 26.7 & 51.2    \\
    \bottomrule
  \end{tabular}
  \caption{Effect of learning rate (LR) and training batch size (batch size) on Bhojpuri-Hindi fine-tuning performance}
  \label{tab:lr_comp}
\end{table}

\begin{table}[th]
  \centering
  \begin{tabular}{lrr}
    \toprule
    \textbf{LS} & \textbf{BLEU} & \textbf{chrF++} \\
    \midrule
    0.0    & 30.9         & 55.3 \\
    0.1  & \textbf{33.8}  & \textbf{56.9} \\
    0.2  & 31.8        & 56.6 \\
    \bottomrule
  \end{tabular}
  \caption{Effect of label smoothing (LS) on Bhojpuri--Hindi fine-tuning performance with LR = 1e-5 and batch size = 10.}
  \label{tab:ls_comp}
\end{table}

\subsection{Data augmentation}
To address the limited training data for the Bhojpuri-Hindi language pair, we implemented two established speech augmentation techniques.

\subsubsection{SpecAugment}
We applied spectrogram masking with time masks (max 30 frames) and frequency masks (max 30 mel-frequency bins), creating diverse variations that forced the model to rely on broader contextual information rather than specific acoustic features.

\subsubsection{Speed perturbation}
We implemented speed factors of 0.9x, 1.0x, and 1.1x to simulate different speaking rates without changing pitch, effectively tripling our training data with realistic variations and improving the robustness of the model to natural speaking rate differences among Bhojpuri speakers.

By combining these complementary methods, we expanded the diversity of training data without requiring additional recordings. The impact on translation quality is shown in Table~\ref{tab:sp}.

\begin{table}[th]
  \centering
  \begin{tabular}{ccr}
    \toprule
    \textbf{SP} & \textbf{SA} & \textbf{BLEU}  \\
    \midrule
    False & False & 31.8 \\
    \textbf{False} & \textbf{True} & \textbf{33.7} \\
    True & False &  32.7 \\
    True & True & 32.4  \\
    \bottomrule
  \end{tabular}
  \caption{Effect of Speed perturb (SP) and  SpecAugment (SA) on Bhojpuri--Hindi fine-tuning performance with LR = 1e-5, batch size = 10, and LS = 0.1}
  \label{tab:sp}
\end{table}


\subsection{Joint-finetuning approach}
We implemented cross-lingual transfer learning by integrating Marathi-Hindi and Bhojpuri-Hindi language pairs into a unified training framework. This approach leverages the linguistic similarities between these Indo-Aryan languages, which share phonological characteristics, lexical resources, and syntactic structures.

Our methodology combined the Marathi-Hindi parallel corpus with the limited Bhojpuri-Hindi dataset during SeamlessM4T model adaptation, thereby expanding the training data while introducing linguistically relevant patterns from Marathi. To mitigate catastrophic forgetting, we implemented sequential fine-tuning with an initial joint training phase followed by Bhojpuri-only fine-tuning. We tested this approach under three conditions: one epoch, two epochs, and training until convergence, as shown in Table~\ref{tab:jointft}.

\begin{table*}[th]
\centering
\begin{tabular}{lrrrrrrrr}
\toprule
\textbf{LR} & \textbf{LS} & \textbf{Batch size} & \textbf{SP} & \textbf{SA} & \textbf{Warm up steps} & \textbf{Patience} & \textbf{Beam size} & \textbf{BLEU} \\
\midrule
1e\_5 & 0.1 & 10 & False & False & 100 & 5 & 5 & 33.1 \\
\midrule
1e\_5 & 0.1 & 10 & False & False & 100 & 5 & 10 & 33.8 \\
1e\_5 & 0.1 & 32 & False & False & 100 & 5 & 10 & 34.0 \\
\midrule
1e\_5 & 0.1 & 32 & False & True & 250 & 5 & 10 & 35.3 \\
\textbf{1e\_5} & \textbf{0.1} & \textbf{32} & \textbf{False} & \textbf{True} & \textbf{250} & \textbf{10} & \textbf{10} & \textbf{36.4} \\
1e\_5 & 0.1 & 32 & False & True & 250 & 20 & 10 & 35.6 \\
\bottomrule
\end{tabular}
\caption{BLEU scores for various hyperparameter configurations during fine-tuning of SeamlessM4T. We varied the learning rate (LR), label smoothing (LS), batch size, speed perturbation (SP), SpecAugment (SA), warm-up steps, early stopping patience, and beam size. The highest BLEU score (36.41) was obtained with SA enabled, patience set to 10, and a beam size of 10.}
\label{tab:final}
\end{table*}

\begin{table}[th]
\centering
\begin{tabular}{lrr}
\toprule
\textbf{Strategy} & \textbf{Epochs} & \textbf{BLEU}  \\
\midrule
Joint finetuning (JF) & Convergence & 34.6 \\
\midrule
 \textbf{JF + monolingual bhoj} &\textbf{1} &  \textbf{36.0} \\
 JF + monolingual bhoj & 2 & 35.6  \\
 JF + monolingual bhoj & Convergence & 35.4 \\
\bottomrule
\end{tabular}
\caption{BLEU scores for contrastive model with the same configuration as highest BLEU in Table~\ref{tab:final} }
\label{tab:jointft}
\end{table}

\section{Results and analysis}
\label{sec:results}
This section presents an analysis of the experimental results obtained from our primary and contrastive models. The \textbf{primary model} encompasses an optimized combination of hyperparameter configuration and data augmentation techniques that yielded the highest BLEU score in our evaluations.

As demonstrated in Table~\ref{tab:sp}, the application of SpecAugment during fine-tuning produced superior performance compared to other augmentation strategies. During inference, we systematically evaluated translation quality across multiple beam search configurations (sizes 1, 5, and 10) to determine the optimal decoding approach.

Table~\ref{tab:final} represents a comprehensive comparison of various experimental configurations and their corresponding performance metrics. Increasing the beam size from 5 to 10 while maintaining all other parameters constant yielded a modest improvement in translation quality. Subsequently, with beam size fixed at 10, enlarging the training batch size from 10 to 32 further enhanced performance by 0.21 BLEU points. The combination of beam size 10, batch size 32, increased warmup steps from 100 to 250, and the introduction of SpecAugment collectively improved the BLEU score from 34.01 to 35.38. Our optimal configuration, which additionally increased early stopping patience from 5 to 10, achieved the highest performance with 36.41 BLEU. Notably, further extending patience to 20 epochs resulted in performance degradation, suggesting potential overfitting.

For our \textbf{contrastive model}, we implemented a multilingual fine-tuning strategy that jointly trained on Marathi and Bhojpuri data using the hyperparameter configuration that previously achieved the highest performance (36.4 BLEU). This multilingual model initially underperformed compared to our monolingual system, potentially due to catastrophic forgetting. To mitigate this issue, we conducted additional fine-tuning on Bhojpuri data exclusively for various epoch counts. 

As shown in Table~\ref{tab:jointft}, performance peaked after a single epoch of Bhojpuri-specific fine-tuning and subsequently declined with additional epochs, suggesting that extended training on previously observed data may lead to overfitting. Consequently, our contrastive submission consisted of the joint Marathi-Bhojpuri model with one additional epoch of Bhojpuri--Hindi exclusive fine-tuning, which produced results comparable to our optimised monolingual configuration.

Table~\ref{tab:testresult} presents the BLEU scores obtained on both test and development datasets. In the IWSLT 2025 shared task \cite{iwslt2025}, the highest reported BLEU score was 10.7, representing a significant decrease compared to IWSLT 2024 \cite{ahmad-etal-2024-findings}, where scores reached approximately 24.4. Our primary and contrastive models achieved BLEU scores of 9.9 and 10.2 respectively on the test set, while demonstrating substantially higher performance on the development set with scores of 36.4 and 36.0. The considerable performance gap between development and test sets suggests potential domain mismatch between the datasets or possible data quality issues in the test set, warranting further investigation.

\begin{table}[th]
\centering
\begin{tabular}{lrr}
\toprule
\textbf{Model} & \textbf{Dev BLEU} & \textbf{Test BLEU}  \\
\midrule
Primary   & 36.4 & 9.9   \\
\midrule
Contrastive & 36.0 &  10.2 \\
\bottomrule
\end{tabular}
\caption{BLEU scores for primary and contrastive models using the same configuration as highest BLEU scores in Table~\ref{tab:final} and Table~\ref{tab:jointft} for both dev and test dataset }
\label{tab:testresult}
\end{table}

\subsection{Error analysis}
Our systematic examination of translation outputs revealed several factors in the development dataset that affected ST performance. Analysis of audio-transcript relationships identified multiple inconsistencies impacting model performance. We observed three key patterns when comparing reference and target word counts: (1) When reference counts exceeded target counts, low BLEU scores often resulted from incomplete audio recordings paired with complete reference transcripts, audio-transcript misalignment, or redundant reference content; (2) Equal word counts with low BLEU scores frequently corresponded with noisy recordings; (3) Cases where target counts exceeded reference counts typically involved recordings with significant acoustic interference.

Numerical content presented particular challenges. We identified inconsistent representation formats (e.g., "8 crores 74 lakhs" in audio versus "87.4 lakhs" in text), incomplete numerical transcription (e.g., in audio, numbers are spoken in English as "Fifteen" whereas in reference text they appear in Hindi as "Pandrah(hindi)"), and instances where equal numerical representation corresponded with degraded audio quality. These findings highlight the importance of audio-transcript alignment and standardized numerical representation in speech translation datasets, particularly for low-resource language evaluation.

\section{Conclusion}
\label{sec:conc}
Our submission to the IWSLT 2025 evaluation campaign for low-resource and dialectal speech translation advances Bhojpuri--Hindi ST through a combination of hyperparameter optimisation, data augmentation, and cross-lingual joint fine-tuning. By leveraging the SeamlessM4T medium model (1.2B parameters) and systematically exploring optimal training configurations, we demonstrate significant performance gains despite the challenges posed by limited parallel data.
Our results show that even established techniques like SpecAugment and speed perturbation, when carefully implemented, can lead to substantial improvements in low-resource speech translation tasks, expanding our effective training data threefold. Additionally, we found that joint training with Marathi—a linguistically related Indo-Aryan language—followed by sequential Bhojpuri-specific adaptation provides an effective strategy to mitigate data sparsity and improve generalisation across diverse speech patterns.

For future work, we plan to explore more sophisticated augmentation techniques such as noise injection, pitch shifting, and cross-speaker synthesis. We also intend to investigate self-supervised pretraining on monolingual Bhojpuri speech data and further extend our cross-lingual approach to additional Indo-Aryan languages. As low-resource ST continues to evolve, we believe that modular, linguistically informed adaptation pipelines will play a key role in advancing the real-world applicability of such systems for under represented language communities.

\section{Acknowledgements}

Santosh Kesiraju was supported by Ministry of Education, Youth and Sports of the Czech Republic (MoE) through the OP JAK project ``Linguistics, Artificial Intelligence and Language and Speech Technologies: from Research to Applications'' (ID:CZ.02.01.01/00/23\_020/0008518).

\bibliography{anthology,custom}
\end{document}